\documentclass[letterpaper, 10 pt, conference]{ieeeconf}  

\IEEEoverridecommandlockouts                              

\usepackage{cite}
\usepackage{amsmath,amssymb,amsfonts}
\usepackage{algorithmic}
\usepackage{graphicx}
\usepackage{textcomp}
\usepackage{xcolor}
\usepackage{tabularx}
\usepackage{multirow}
\usepackage{subcaption}
\usepackage[capitalize]{cleveref}
\usepackage{flushend}
\def\BibTeX{{\rm B\kern-.05em{\sc i\kern-.025em b}\kern-.08em
    T\kern-.1667em\lower.7ex\hbox{E}\kern-.125emX}}

\begin{document}

\title{ReefGlider: A Highly Maneuverable Vectored Buoyancy Engine Based Underwater Robot\\
{}
}

\author{Kevin Macauley$^{1,2}$, Levi Cai$^{3}$, Peter Adamczyk$^2$, and Yogesh Girdhar$^1$
\thanks{*This work was supported in part by NSF Grant 2133029, and the Investment in Science Fund at WHOI.}
\thanks{$^1$Kevin Macauley (as a guest student) and Yogesh Girdhar are with the Woods Hole Oceanographic Institution (WHOI) {\tt\small yogi@whoi.edu}}
\thanks{$^2$Kevin Macauley and Peter Adamczyk are with the University of Wisconsin-Madison {\tt\small \{kcmacauley, peter.adamczyk\}@wisc.edu}} \thanks{$^3$ Levi Cai is with the Massachusetts Institute of Technology and Woods Hole Oceanographic Institution Joint Program {\tt\small cail@mit.edu}}}



\maketitle

\begin{abstract}
There exists a capability gap in the design of currently available autonomous underwater vehicles (AUV). Most AUVs use a set of thrusters, and optionally control surfaces, to control their depth and pose. AUVs utilizing thrusters can be highly maneuverable, making them well-suited to operate in complex environments such as in close-proximity to coral reefs. However, they are inherently power-inefficient and produce significant noise and disturbance. Underwater gliders, on the other hand, use changes in buoyancy and center of mass, in combination with a control surface to move around. They are extremely power efficient but not very maneuverable. Gliders are designed for long-range missions that do not require precision maneuvering. Furthermore, since gliders only activate the buoyancy engine for small time intervals, they do not disturb the environment and can also be used for passive acoustic observations. In this paper we present ReefGlider, a novel AUV that uses only buoyancy for control but is still highly maneuverable from additional buoyancy control devices. ReefGlider bridges the gap between the capabilities of thruster-driven AUVs and gliders. These combined characteristics make ReefGlider ideal for tasks such as long-term visual and acoustic monitoring of coral reefs. We present the overall design and implementation of the system, as well as provide analysis of some of its capabilities.
\end{abstract}


\section{Introduction}

There is a growing need for underwater robotic platforms to provide monitoring and exploration capabilities in complex and dynamic environments, such as coral reefs or kelp forests \cite{llewellyn_getting_2015}. Specifically, we are interested in robotic platforms that can provide high-fidelity observations using sensors such as hydrophones and cameras, while minimizing noise generated from the vehicle and maximizing observation time. These vehicles need to be (1) low-energy, (2) minimally disturbant to the environment, and (3) highly maneuverable. 

Underwater gliders \cite{rudnick_underwater_2004, cauchy2023gliders, lembke2018using}, wavegliders \cite{manley_wave_2010}, and sailing drones \cite{gentemann_saildrone_2020, de_robertis_long-term_2019, mordy_advances_2017} are in active use for these purposes, but are often used for large-scale oceanography and used to resolve environmental features on the order of tens of meters to kilometers. Thruster-actuated robots such as CUREE \cite{Girdhar2023} have also been used for passive acoustic monitoring of coral reefs to observe acoustic bioactivity. To deal with thruster noise, CUREE interleaves its trajectory with periods of drifting during which it has no control. The approach however has problems because acoustic samples can be collected over only very short time periods when the thrusters are off. Furthermore, due to its continuous thruster use for most of its operating time, it has a very short operating time (2-3 hours) and is not suitable for long-term monitoring tasks.

In this paper, we propose a novel Vectored Buoyancy Control (VBC) system that enables a new class of AUVs that are energy efficient, highly maneuverable, and low-noise. ReefGlider utilizes VBC to provide 6 degree-of-freedom motion, though non-holonomic control, at sub-meter resolutions.

In the remainder of the paper we provide background \cref{sec:background}, the hardware design of the system \cref{sec:hardware}, an initial control implementation \cref{sec:control}, real-world experiments using the entire platform \cref{sec:experiments}, and finally a discussion and concluding thoughts in \cref{sec:conclusion}.
\begin{figure}
    \centering
    \includegraphics[width=\linewidth]{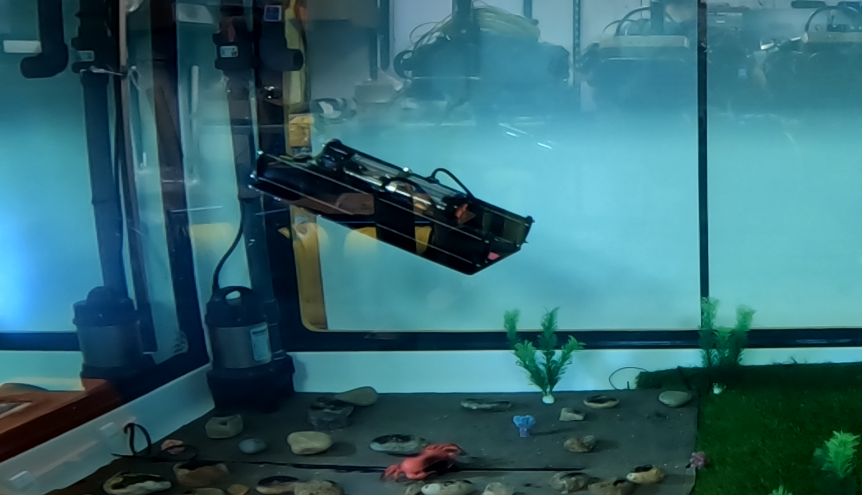}
    \caption{ReefGlider show here in a tank is a highly maneuverable underwater robot that uses buoyancy for propulsion, but unlike other gliders is highly maneuverable.}
    \label{fig:vbcd-glamour}
\end{figure}
\section{Background}
\label{sec:background}
\subsection{Buoyancy control for underwater vehicles}
Buoyancy control has long been used as a method for low-cost, low-disturbance, and low-energy actuation for underwater vehicles. Underwater gliders are perhaps the most widely used system that utilizes this strategy, relying almost exclusively on buoyancy and barycentric control. Vehicles such as UW Seagliders, Spray gliders, or Slocum gliders are commonly used to provide long-term monitoring of large-scale ocean regions. They are especially effective for these tasks because they are low-energy, so their mission lengths can be up to several months \cite{rudnick_underwater_2004, sherman2001autonomous, carneiro_using_2019}. 

Additionally, in most cases, gliders only need to \textit{occasionally} move a mass weight, or modify buoyancy via a bladder or linear actuator, making them relatively silent, allowing their use for acoustic measurements \cite{jiang_use_2019}.

Underwater gliders typically use wings (either passive or active) and a movable mass to control orientation, but this results in large turning radii, such as the 7m radius of the Slocum or larger for others \cite{page_highly_2017,webb_slocum_2001}. In turn, this makes them mostly appropriate for large-scale data collection, often on the order of kilometers \cite{rudnick_underwater_2004}. Some gliders, and most highly maneuverable vehicles, such as \cite{bhat_hydrobatics_2018, bhat_hydrobatics_nodate, duecker_hippocampusx_2020}, are designed to include thrusters and additional wing surfaces for added maneuverability, but thrusters nullify many of the silent and energy-efficient benefits of the buoyancy-driven vehicle. Osse et al. \cite{osse2018oculus} used a hydraulic amplifier to allow faster buoyancy changes for shallow water operation. Page et al. \cite{page_highly_2017} introduced an additional roll mechanism for the mass, which was used on a smaller glider to reduce its turning radius to 3-meters while maintaining the other desired characteristics. 

Here, we propose an alternative strategy, using additional buoyancy control engines as control actuators rather than thrusters or masses. This enables similar energy and noise savings, while allowing for much higher-resolution, sub-meter control, including zero turning radius. We term this type of actuation \textit{Vectored Buoyancy Control} (VBC) .

\subsection{Buoyancy control mechanisms}

Buoyancy control systems are typically designed around a \textit{single} mechanism to control the buoyancy of the vehicle. These are either electro-hydraulically driven, typically involving pumping oil to and from an external and internal bladder, or electro-mechanically driven \cite{carneiro_using_2019}, usually consisting of a single linear actuator that moves a barrier to change a housing volume and displace water. In this paper, we focus primarily on an electro-mechanical system, that instead utilizes several linear actuators changing the volume of several housings, due to its design simplicity and energy efficiency as in \cite{carneiro_using_2019,falcao_carneiro_design_2022}, though an electro-hydraulic solution can be used well.

 We hypothesize that these types of systems can enable new sampling strategies for vehicles operating in more dynamic environments requiring higher-resolution spatiotemporal data gathering, such as coral reefs and kelp forests, while also reducing noise and energy-consumption.

\section{System Design}
\label{sec:hardware}

\begin{figure}
    \centering
    \includegraphics[width=1\linewidth]{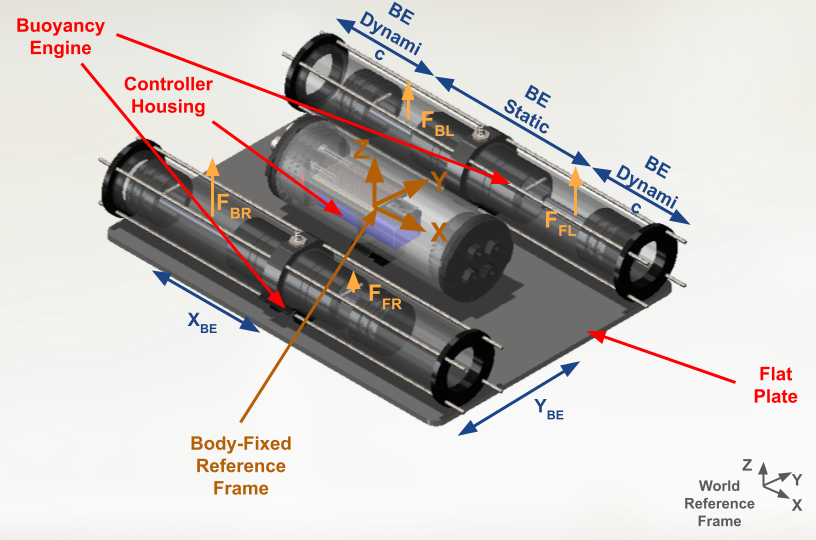}
    \caption{Isometric view of the ReefGlider and relevant frames of reference.}
    \label{fig:iso}
\end{figure}

\begin{figure}
    \centering
    \includegraphics[width=1\linewidth]{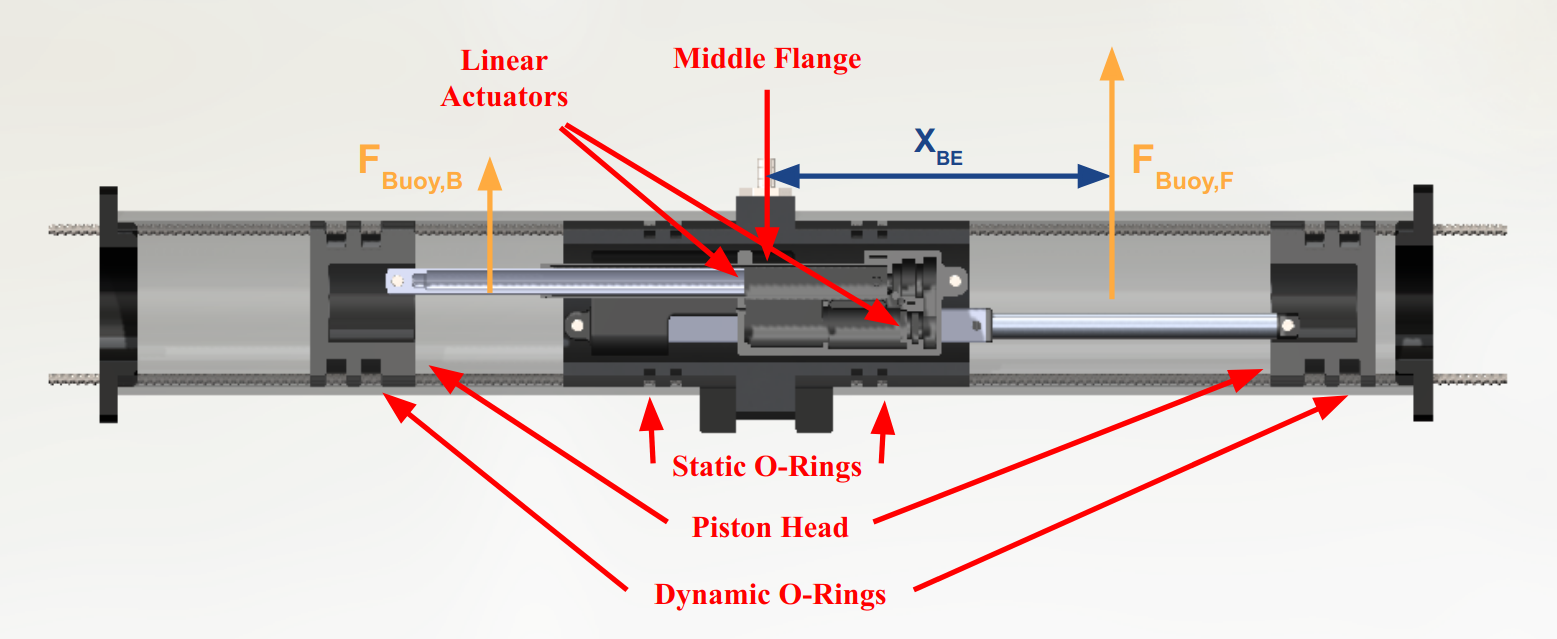}
    \caption{Section-view of a single Buoyancy Engine.}
    \label{fig:section_view_cad}
\end{figure}

    \subsection{Overview}
    The ReefGlider is designed to be able to change roll, pitch, and overall buoyancy without \textit{requiring} long transects as many underwater gliders. At a high-level, the VBC shifts the center of buoyancy location and buoyancy force magnitude in order to control orientation and vertical motion. A passive component, such as a planar piece or wing, creates drag which can be used to enable lateral motions. The ReefGlider achieves this by controlling 4 individual buoyancy control mechanisms, which work by changing the amount of water displaced by four linear pistons. The pose is controlled by independently moving the pistons to create a difference between the center of buoyancy (COB) and the center of mass (COM) shown in \cref{fig:iso}. While this is similar to traditional underwater gliders, we use a single style of actuation to control both the COM and COB, rather than a separate style of mass and buoyancy actuators, which also enables finer control over the forces acting on the vehicle.

    \subsection{Hardware details}
    The overall design of the ReefGlider's VBC system consists of two buoyancy engines (each contains two linear actuators, resulting in four total actuators) and a controller housing mounted to a passive flat plate as shown in \cref{fig:iso}. We note that the passive component can be easily modified to achieve different dynamic goals. The buoyancy engines contain two independent pistons of which are displaced by Actuonix P16-100-256-12-P Linear Actuators \cite{noauthor_actuonix_nodate}. The middle flange was designed such that two linear actuators could fit side by side as shown in \cref{fig:section_view_cad}. This reduces the static volume of the buoyancy engine (BE static) and the overall footprint of the device. 

    A Teensy 4 microcontroller is the main control module responsible for reading data from sensors and actuating the independent linear actuators. The length of actuation of each linear actuator is commanded by a PWM signal from the Teensy to the external RC control board which is responsible for the position control of the linear actuators. To provide orientation feedback, the ReefGlider relies on an inertial measurement unit (IMU), where we use the Bosch BNO055. Similarly, depth or pressure feedback is provided by a BlueRobotics Bar30 pressure sensor. Power is supplied by a 3-S LiPo battery at 11.1 V directly to the linear actuators and all other electronics through a Pololu 5V voltage regulator.

    To meet the tight tolerance necessary for o-rings, the middle flange, piston head, and outer tubes were custom machined. The middle flange was machined from aluminum so that a simple cable penetrator could be threaded into it and anodized for corrosion resistance. The piston head and outer tube were machined out of acetal to reduce machining costs while still being corrosion resistant. All o-ring dimensions were determined using the Parker O-ring Handbook \cite{parker1992parker}. Four threaded rods and two 3D-printed end caps are used to axially constrain the buoyancy engines.  A 3D-printed bracket is fastened to both buoyancy engines and to the passive planar component, and hence to each other.

    All mechanical dimensions and details are shown in \cref{tab:geometry_mass} and \cref{fig:iso}.

\section{Modeling and Control Architecture}
\label{sec:control}
A simplified linearized physics model of the ReefGlider was created using cylinders and boxes to estimate the geometry of the vehicle. The controller housing is modeled as a cylinder with its COM and COB in the center of the cylinder. The plate is modeled as box with its own COM and COB. Each buoyancy engine is modeled as three independent cylinders with their own COM and COB: two dynamic cylinders and one static cylinder as shown in \cref{fig:section_view_cad}. The static cylinder represents the volume of water always displaced by the buoyancy engine an outer diameter of 63.5mm. The length of the dynamic cylinders can vary between 0mm and 100mm, which corresponds to the actuation length of the linear actuator with an outer diameter of 57.15mm. The full vehicle dynamics are shown in \cref{fig:iso}.

\begin{table}[htbp]
    \centering
    \caption{Geometry and mass properties for ReefGlider, corresponding frames of reference are shown in \cref{fig:iso}.}
    \begin{tabularx}{\linewidth}{|c|X|c|X|}
        \hline
        \multirow{2}{*}{Geometry} & \multirow{2}{*}{Dimensions (m)} & \multirow{2}{*}{Mass (kg)} &  Center of Mass \\ & & & x, y, z (m) \\
        \hline
        \multirow{3}{*}{Flat Plate} & Width: 0.330 & \multirow{3}{*}{0.945} & \multirow{3}{*}{0, 0, 0} \\
        & Length: 0.457 & & \\
        & Thickness: 0.006 & & \\
        \hline
        \multirow{2}{*}{BE Static} & Diameter: 0.064 & \multirow{2}{*}{1.505} & \multirow{2}{*}{0, $\pm$0.124, 0.044} \\
        & Length: 0.257 & & \\
        \hline
        \multirow{2}{*}{BE Dynamic} & Diameter: 0.057 & \multirow{2}{*}{(Varies)} & $\pm$(0.127+length), \\
        & Length: 0 to 0.1 & &  $\pm$0.124, 0.044 \\
        \hline
        \multirow{2}{*}{Control Housing} & Diameter: 0.089 & \multirow{2}{*}{1.348} & \multirow{2}{*}{-0.02, 0, 0.025} \\
        & Length: 0.245 & & \\
        \hline
    \end{tabularx}
    \label{tab:geometry_mass}
\end{table}

\subsection{Linearized Open-Loop Control Design}
\label{sec:openloop}
With the defined geometry, an open-loop controller is designed to enable control across pitch, roll, and depth (pressure). Buoyancy force and gravitational forces are the only two forces taken into account in this model. Additionally, the COM and COB are assumed to coincide at the centroid of the simple shapes defined in the physical model previously described. These are defined relative to an inertial frame and the we simplify modelling by assuming the vehicle is operating near a hover state (roll = $0^{\circ}$  and pitch = $0^{\circ}$), creating a linearized model. Using these assumptions a static open loop controller $u_{OL}$ can be developed.

\begin{equation}
    \label{eqn:controller}
    u_{OL}=B^{-1}(\tau_{des}-g_0)
\end{equation}

where

\begin{equation*}
   \tau_{des} = \begin{bmatrix}
F_x\\
F_y\\
F_z\\
\tau_x\\
\tau_y\\
\tau_z\\
\end{bmatrix}, 
u_{OL} =\begin{bmatrix}
x_{FL}\\
x_{BL}\\
x_{BR}\\
x_{FR}\\
\end{bmatrix}, g_0 =\begin{bmatrix}
0\\
0\\
F_{buoy} - gm_{total}\\
0\\
0\\
0\\
\end{bmatrix}
\end{equation*}

\begin{align}
    B &= \begin{bmatrix}
0 & 0 & 0 & 0\\
0 & 0 & 0 & 0\\
1 & 1 & 1 & 1\\
y_{BE} & y_{BE} & -y_{BE} & -y_{BE}\\
x_{BE} & x_{BE} & -x_{BE} & -x_{BE}\\
0 & 0 & 0 & 0\\
 \end{bmatrix} * \alpha \\
 \alpha &= \rho g A_{piston}
\end{align} 

with $F_{i}$ and $\tau_{i}$ as respective forces and torques in the inertial frame described in \cref{fig:iso}, $F_{buoy}$ the static buoyancy, $x_{\{FL, BL, BR, FR\}}$ which are offsets of the corresponding linear actuators, $x_{BE}$ and $y_{BE}$ are static offsets of buoyancy for each cylinder from the body-fixed reference frame shown in \cref{fig:iso}, environmental restoring force $g_0$ which is dependent on the static buoyancy, gravity $g$, and total mass of the vehicle $m_{total}$. In order to further linearize the control calculation, we assume that the change in $\tau_y$ is linearly dependent on the linear actuator offsets, though in actuality it is dependent on its square since both the COB and the buoyancy itself are simultaneously changing. Because $B$ is non-invertible, we use the pseudo-inverse to compute the open-loop control in \cref{eqn:controller}. Also note that by setting $\tau_{des} = 0$, $u_{OL}$ can be treated as a static ballast and trim term, which we do in the proceeding experiments. Because these assumptions introduce errors into the true solution, we design a feedback controller described in the following section.

\subsection{Feedback Control Design}
\label{sec:feedback}

\begin{figure}
    \centering
    \includegraphics[width=1\linewidth]{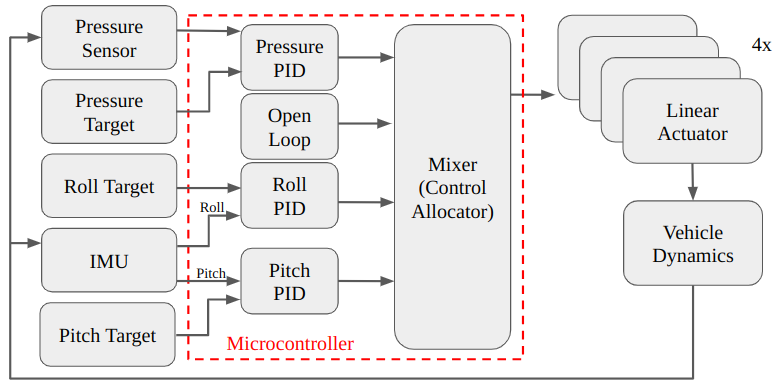}
    \caption{Feedback control diagram for the VBC. The mixer sums the pressure, roll, pitch and Open Loop (OL) control signals and sends the corresponding actuation to each of the linear actuators.}
    \label{fig:feedback}
\end{figure}

To achieve precise control we design feedback controllers for each of depth, pitch, and roll orientations. We utilize separate PID control terms for each parameter, where the feedforward term is calculated using the open-loop control described in \cref{sec:openloop}. Feedback is provided from a pressure sensor for depth regulation and an IMU for pitch and roll orientation, hardware details are provided in \cref{sec:hardware}.  These four control signals were combined in the mixer shown in \cref{eqn:mixer} and produce a position command for each linear actuator. PID control parameters (three for each parameter for a total of nine) were hand-tuned. The overall architecture is shown in \cref{fig:feedback}.

\begin{align}
\label{eqn:mixer}
    u_{mixer} &= \begin{bmatrix}
x_{FL}\\
x_{BL}\\
x_{BR}\\
x_{FR}\\
\end{bmatrix} =  \begin{bmatrix}
1 & -1 & 1 \\
1 & -1 & -1 \\
1 & 1 & -1 \\
1 & 1 & 1 
\end{bmatrix} *  \begin{bmatrix}
PID_{pres.} \\
PID_{roll} \\
PID_{pitch} 
\end{bmatrix}  \\
u_{total} &= u_{OL} + u_{mixer}
\end{align}

\section{Experimental Results}
\label{sec:experiments}
ReefGlider has 4 actuators, and hence it is impossible for it to have instantaneous 6 degrees-of-freedom (DOF) control. However, we demonstrate that the robot can be controlled with full 6 DOF using non-holonomic maneuvers.

Four experiments were conducted to show the maneuverability of the ReefGlider. The first experiment demonstrates the ability of the robot to stay at depth set points while maintaining a "hover" position (approximately zero pitch and roll). The next two experiments show the linear translation capabilities of the robot first by following a sawtooth profile in the pitch direction and second in the roll direction. As in traditional underwater gliders, the ReefGlider is non-holonomic and must combine changing the angle of attack (its orientation) followed by a change a depth in order to move laterally, hence the sawtooth profiles. Finally, we show the ability to control yaw, by performing pitch, roll, pitch maneuvers in series, which illustrate the high maneuverability of this vehicle. These experiments were run completely autonomously by using pre-programmed trajectories that were manually generated, consisting of time-dependent set points in depth, roll, and pitch.

All experiments were performed in a small test tank (roughly 1.5x1.5x3 meters). The depth-hold and yaw experiments included the use of an AprilTag \cite{wang2016iros} mounted on the bottom of the vehicle and measured by an upward-facing camera, in this case a GoPro inside an underwater housing, to get ground-truth measurements.

\subsection{Depth-hold}
\begin{figure}
    \centering

    \begin{subfigure}[b]{0.9\linewidth}
       \includegraphics[width=1\linewidth]{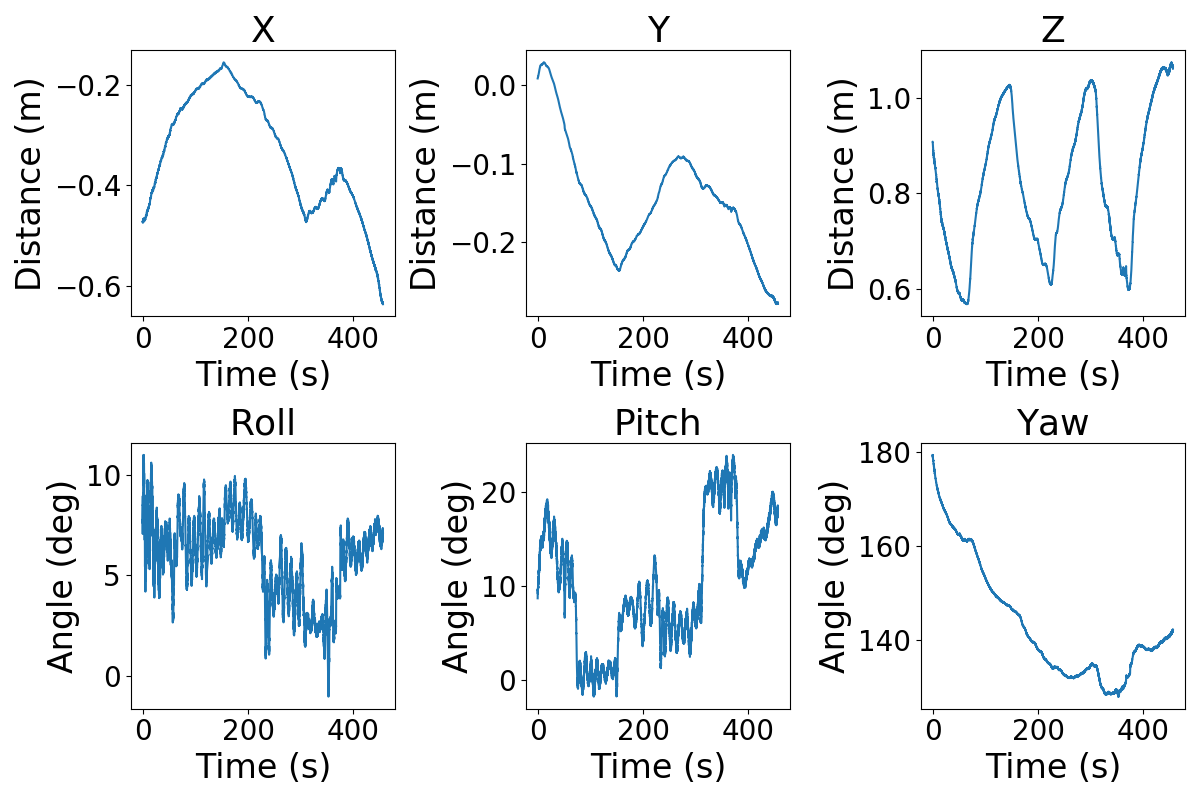}
       \caption{}
        \label{fig:pressure-hold}
    \end{subfigure}
    \begin{subfigure}[b]{0.9\linewidth}
       \includegraphics[width=1\linewidth]{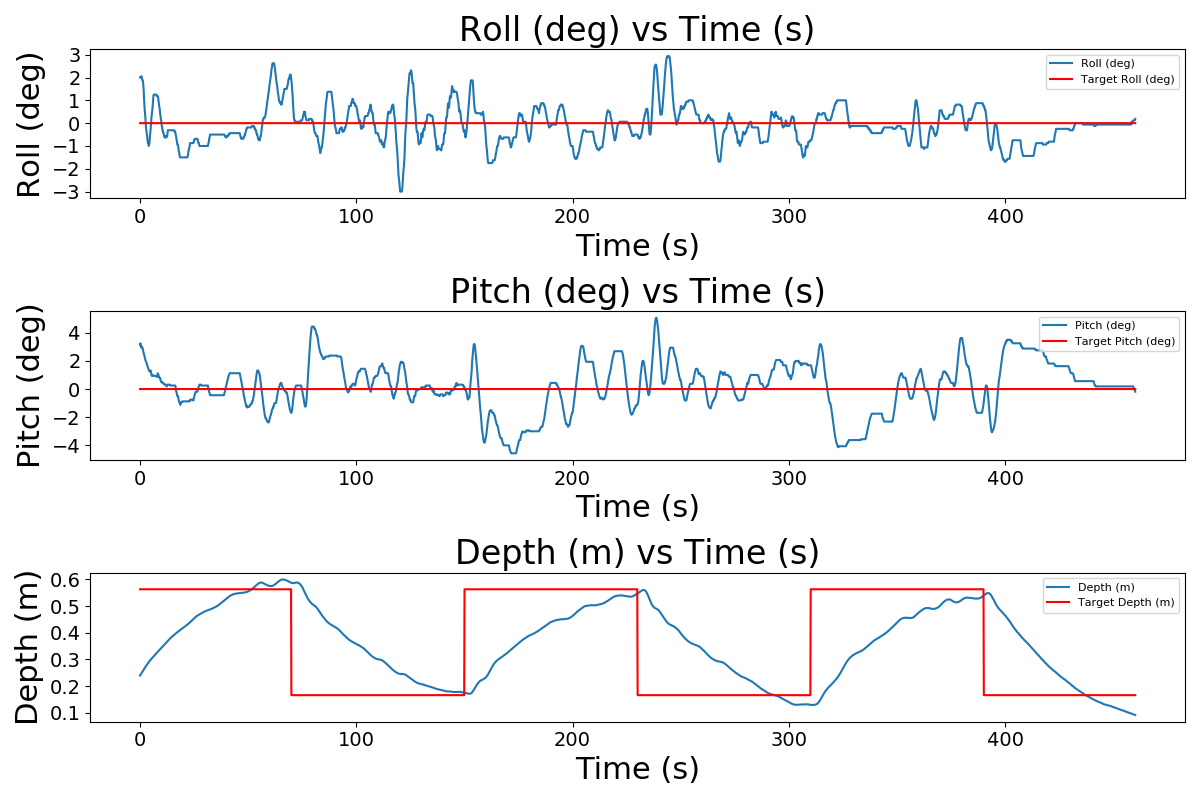}
       \caption{}
        \label{fig:pressure-sensor-data}
    \end{subfigure}
    
    \caption{(a) Depth-hold experiment: ground-truth pose data from AprilTag tracking by a GoPro submerged in test tank. (b) Depth-hold experiment: roll, pitch, and depth readings (in blue) with their corresponding set points plotted (in red) for changing depth set points.}
    \label{fig:pressure}
\end{figure}

The ability for ReefGlider to reach desired depth set points while maintaining zero pitch and zero roll is shown in \cref{fig:pressure-hold} and \cref{fig:pressure-sensor-data}. The depth set points are controlled by their calculated corresponding hydrostatic pressures and are varied between 0.17m (103000 Pa) and 0.58m (107000 Pa). The robot is given a target depth, then allowed to settle before moving onto the next set point. This is shown in \cref{fig:pressure-hold} which shows pose data from the AprilTag measurements, and in \cref{fig:pressure-sensor-data} which plots data recorded from the on-board IMU and pressure sensor. We compute depths in post-processing (though it is possible to do them in real-time), by using the hydrostatic pressure equations, utilizing the density of freshwater for the test tank.

\subsection{Following 3D trajectories via sawtooth profiles}
Similar to a traditional underwater glider, the ReefGlider can achieve lateral motions by changing its angle of attack (due to the passive plate or wing) followed by a change in depth. The primary difference with the ReefGlider is that it is able to not only go forwards \textit{and backwards}, but can also perform these maneuvers along the pitch (inertial x) and roll (inertial y) directions \textit{and in-between} as well.

\begin{figure}[ht!]
    \centering
    \begin{subfigure}[b]{\linewidth}
       \includegraphics[width=1\linewidth]{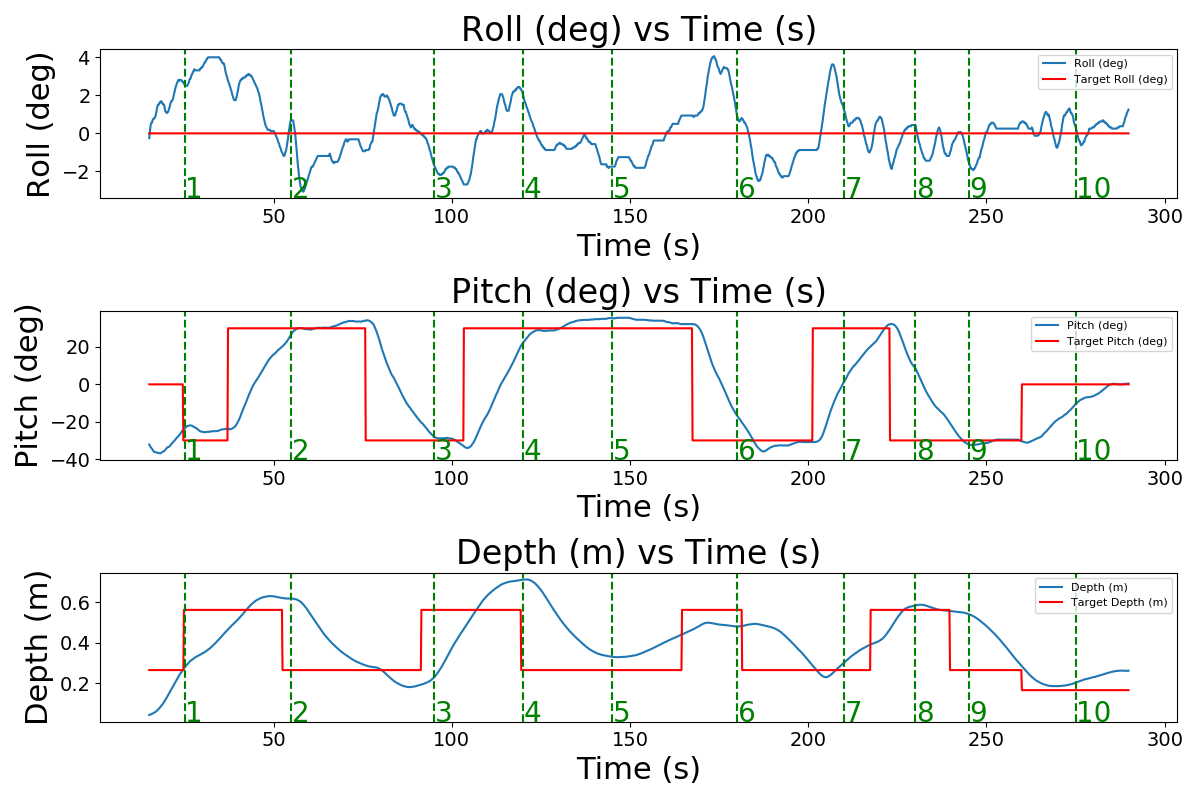}
       \caption{}
    \end{subfigure}
    \begin{subfigure}[b]{0.4\textwidth}
       \includegraphics[width=1\linewidth]{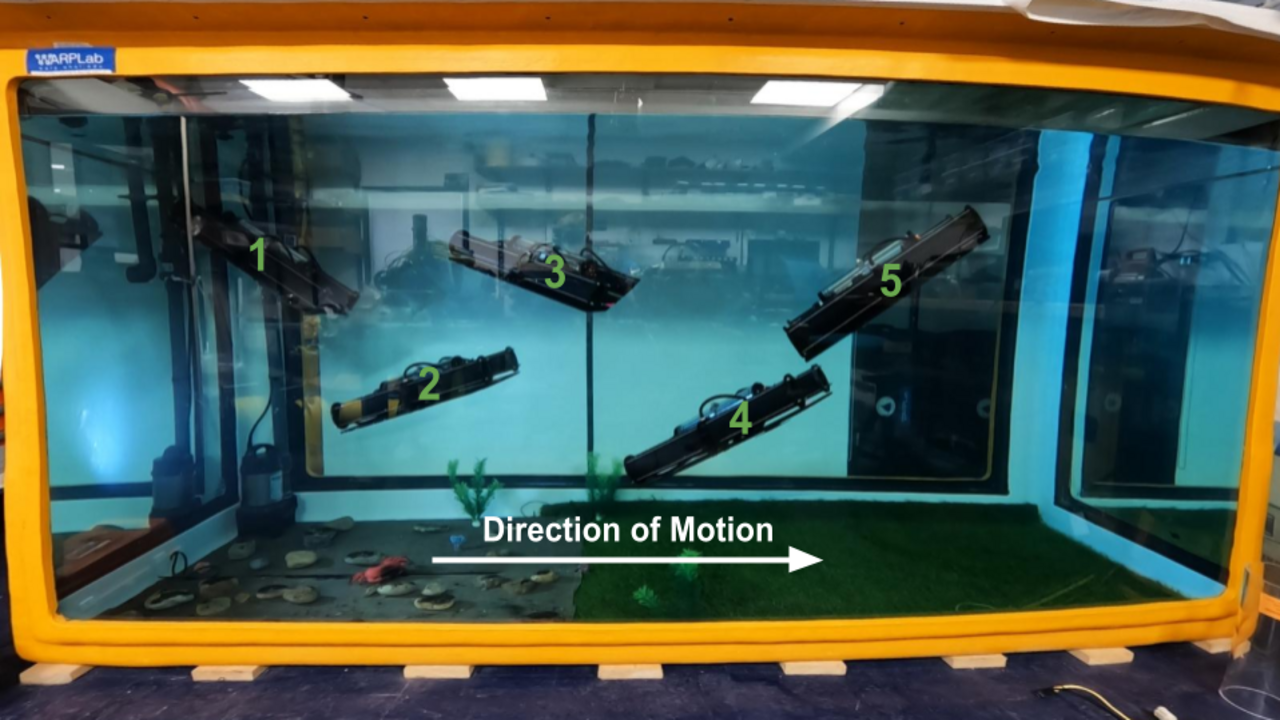}
       \caption{}
    \end{subfigure}
    \begin{subfigure}[b]{0.4\textwidth}
       \includegraphics[width=1\linewidth]{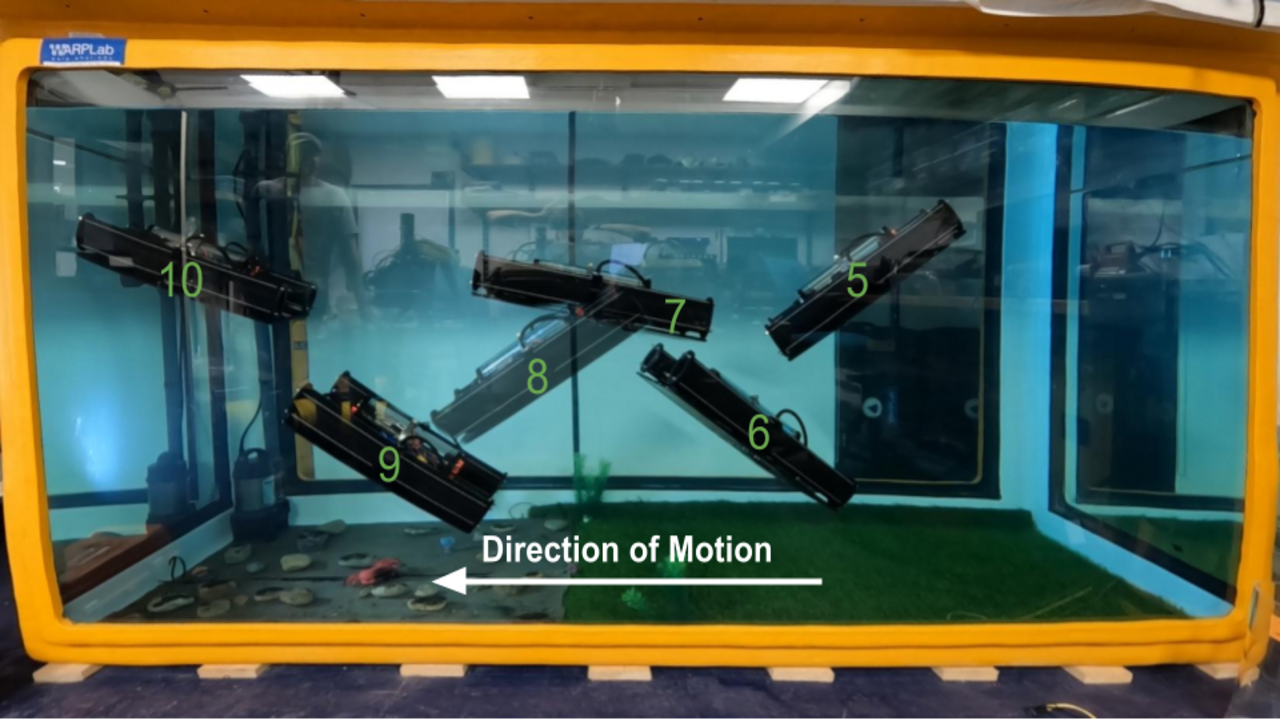}
       \caption{}
    \end{subfigure}
    
    \caption{Pitch-oriented sawtooth transect experiment. In (a) roll, pitch and depth readings (in blue) are given with their corresponding set points plotted (in red) with the vertical dashed green lines corresponding to the positions in (b) and (c). (b) The trajectory of the ReefGlider as it moves from left to right in the test tank merged into a single image. (c) Similarly, the trajectory of the ReefGlider as it moves from right to left as it returns to the starting position.}
    \label{fig:pitch-sawtooth}
\end{figure}

To demonstrate the maneuverability of the robot through the water we perform two separate experiments. First by transiting forward and backward along its inertial x-direction, and then second transiting forward and backward along its inertial y-direction, both using sawtooth profiles. In the first sawtooth experiment, the device is first set to a desired pitch angle, then the overall buoyancy is changed. The induced drag on the robot propels it forward. If this is repeated for different pitch angle and depth targets, a sawtooth pattern can be created as shown in \cref{fig:pitch-sawtooth}. 

\begin{figure}[ht!]
    \centering
    \begin{subfigure}[b]{\linewidth}
       \includegraphics[width=1\linewidth]{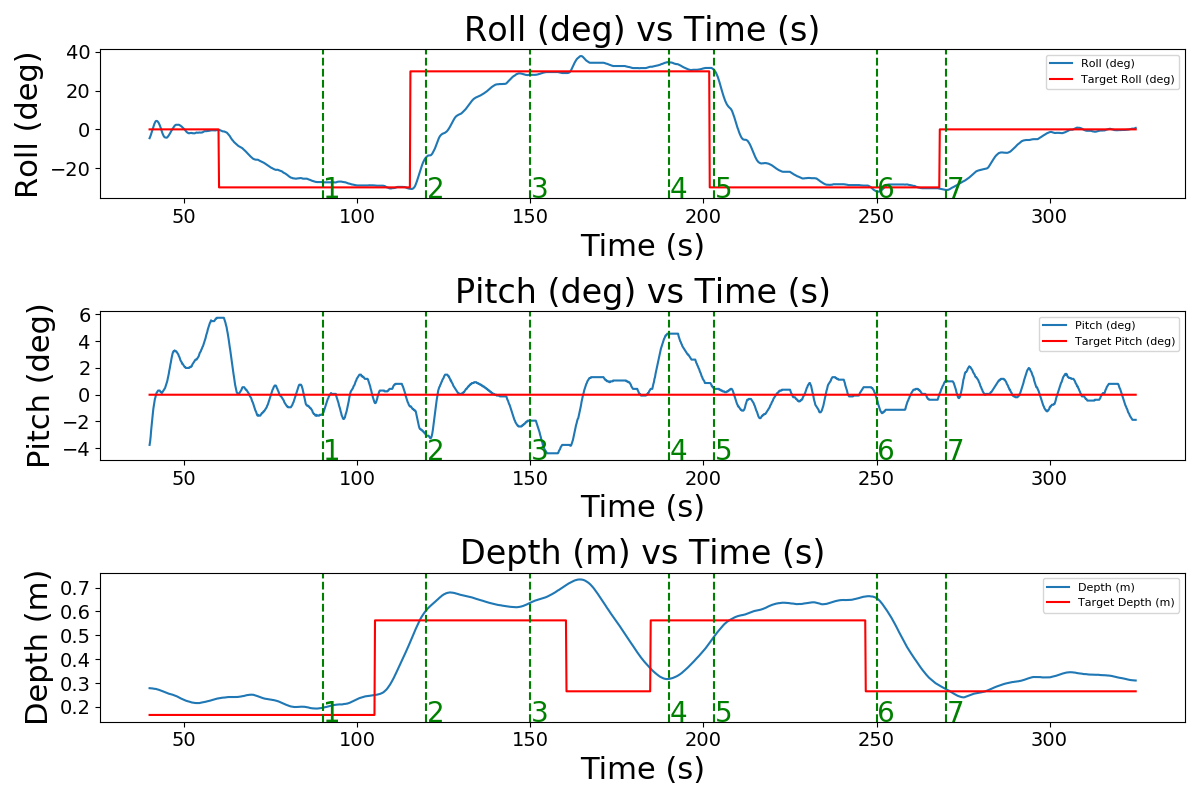}
       \caption{} 
    \end{subfigure}
    \begin{subfigure}[b]{0.4\textwidth}
       \includegraphics[width=1\linewidth]{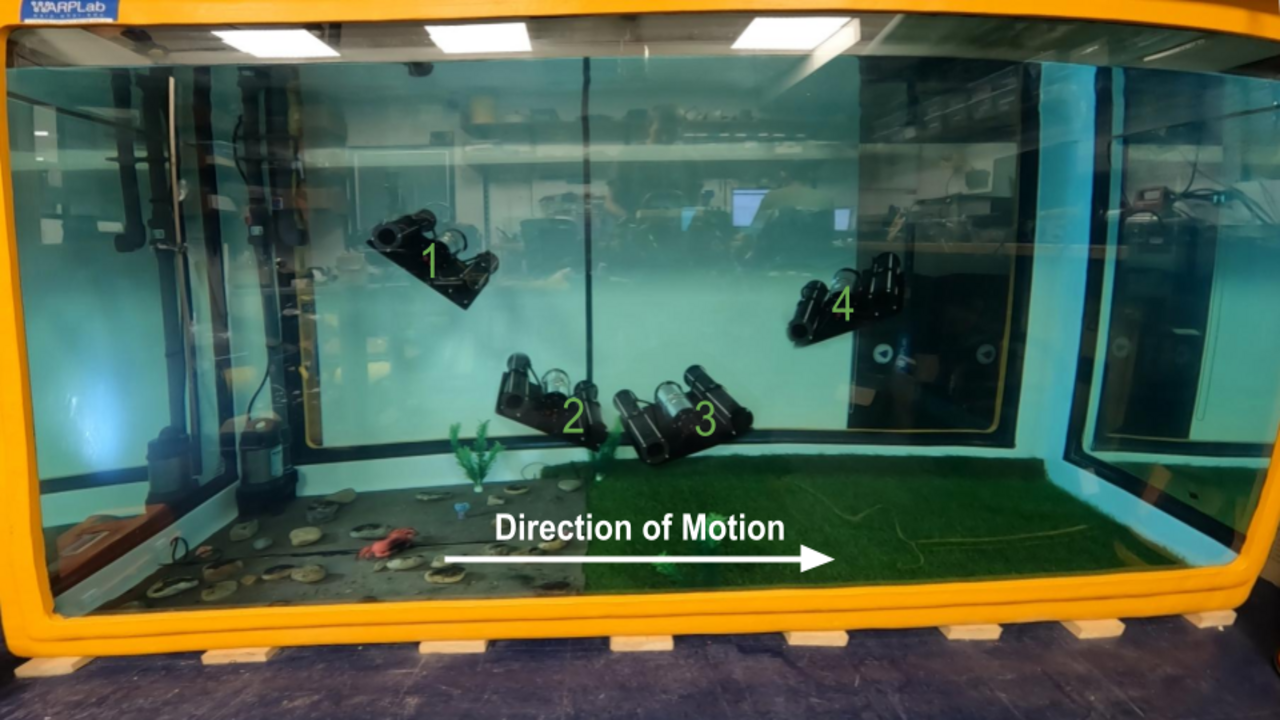}
       \caption{}
    \end{subfigure}
    \begin{subfigure}[b]{0.4\textwidth}
       \includegraphics[width=1\linewidth]{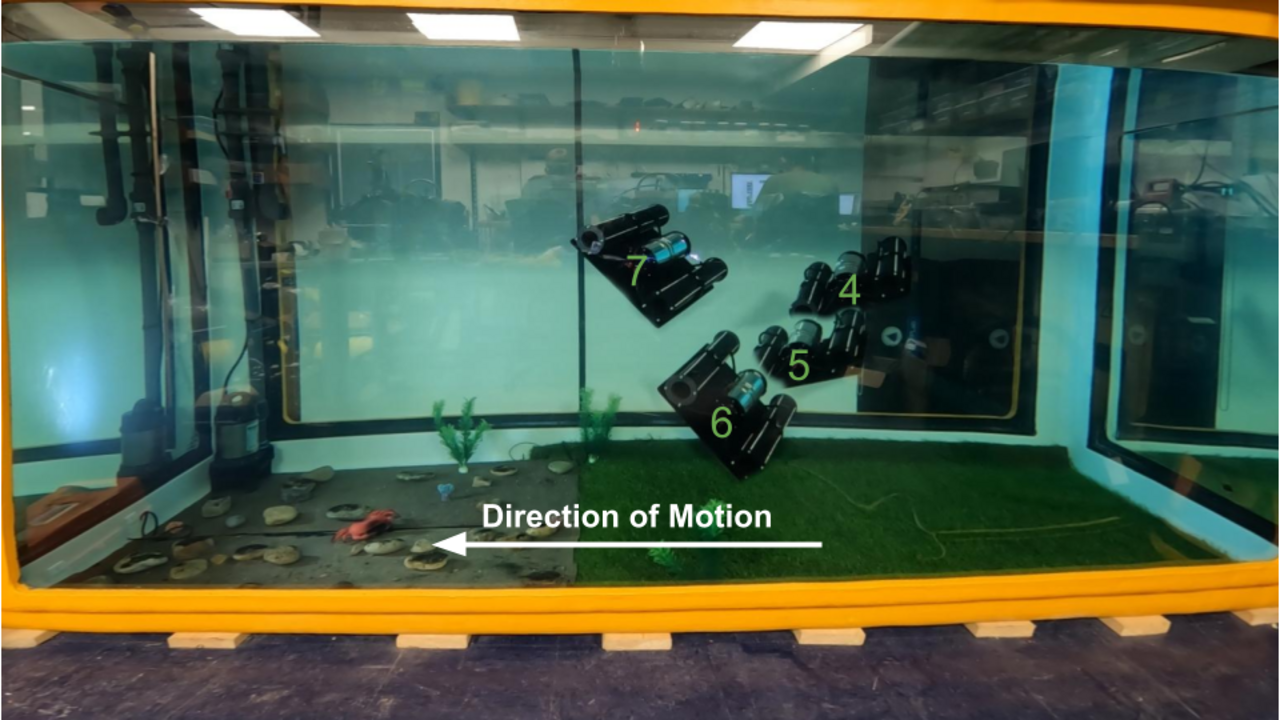}
       \caption{}
    \end{subfigure}
    
    \caption{Roll-oriented sawtooth transect experiment. In (a) roll, pitch and depth readings (in blue) with their corresponding set points plotted (in red) with the vertical dashed green lines corresponding to the positions in (b) and (c). (b) The trajectory of the ReefGlider as it moves from left to right in the test tank merged into one photo. (c) Similarly, the trajectory of the ReefGlider as it moves from right to left as it returns to the starting position.}
    \label{fig:roll-sawtooth}
\end{figure}

Similarly, the robot can move laterally in the y-direction as shown in \cref{fig:roll-sawtooth}. This demonstrates the robots ability to change directions with zero turning radius. While it is non-holonomic and still requires sawtooth patterns, they can be performed faster, allowing the vehicle to visit and station-keep near more complex benthic features or dynamic regions.

\subsection{Yaw experiment}
\begin{figure}
    \centering

    \begin{subfigure}[b]{\linewidth}
       \includegraphics[width=\linewidth]{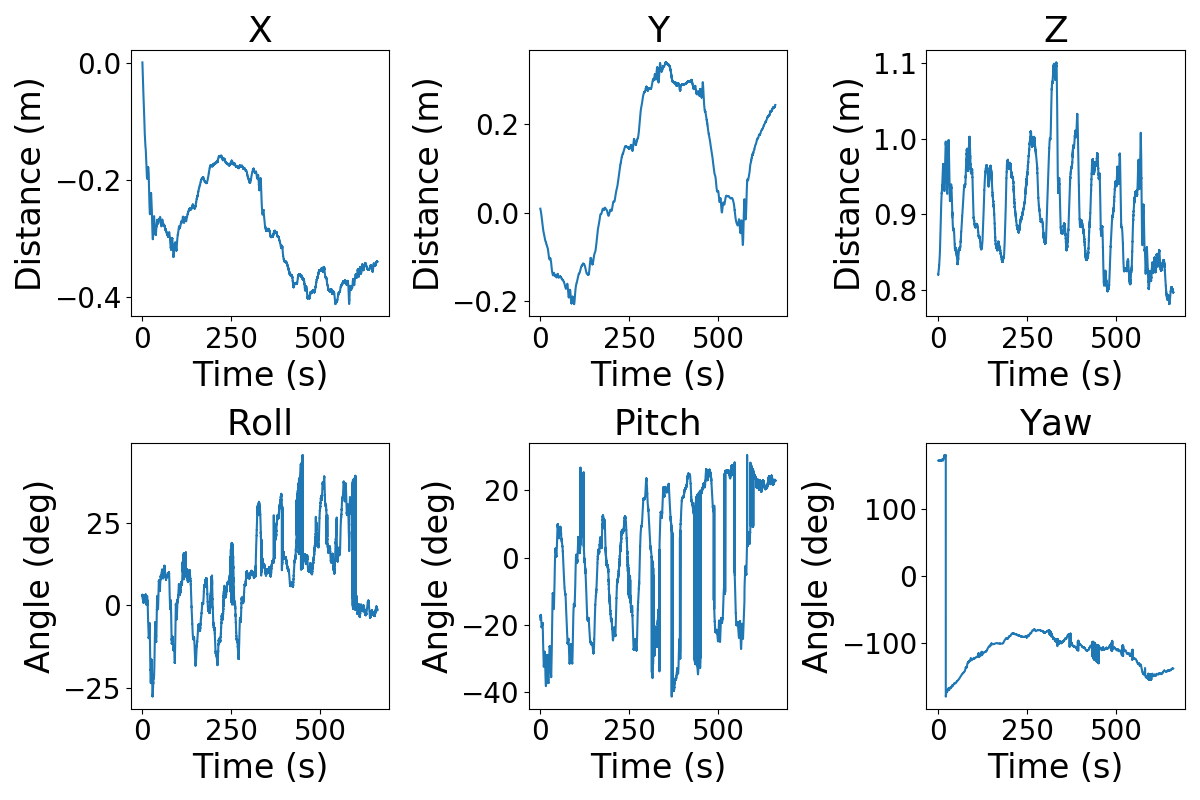}
       \caption{}
        \label{fig:yaw-pose}
    \end{subfigure}
    \begin{subfigure}[b]{\linewidth}
       \includegraphics[width=\linewidth]{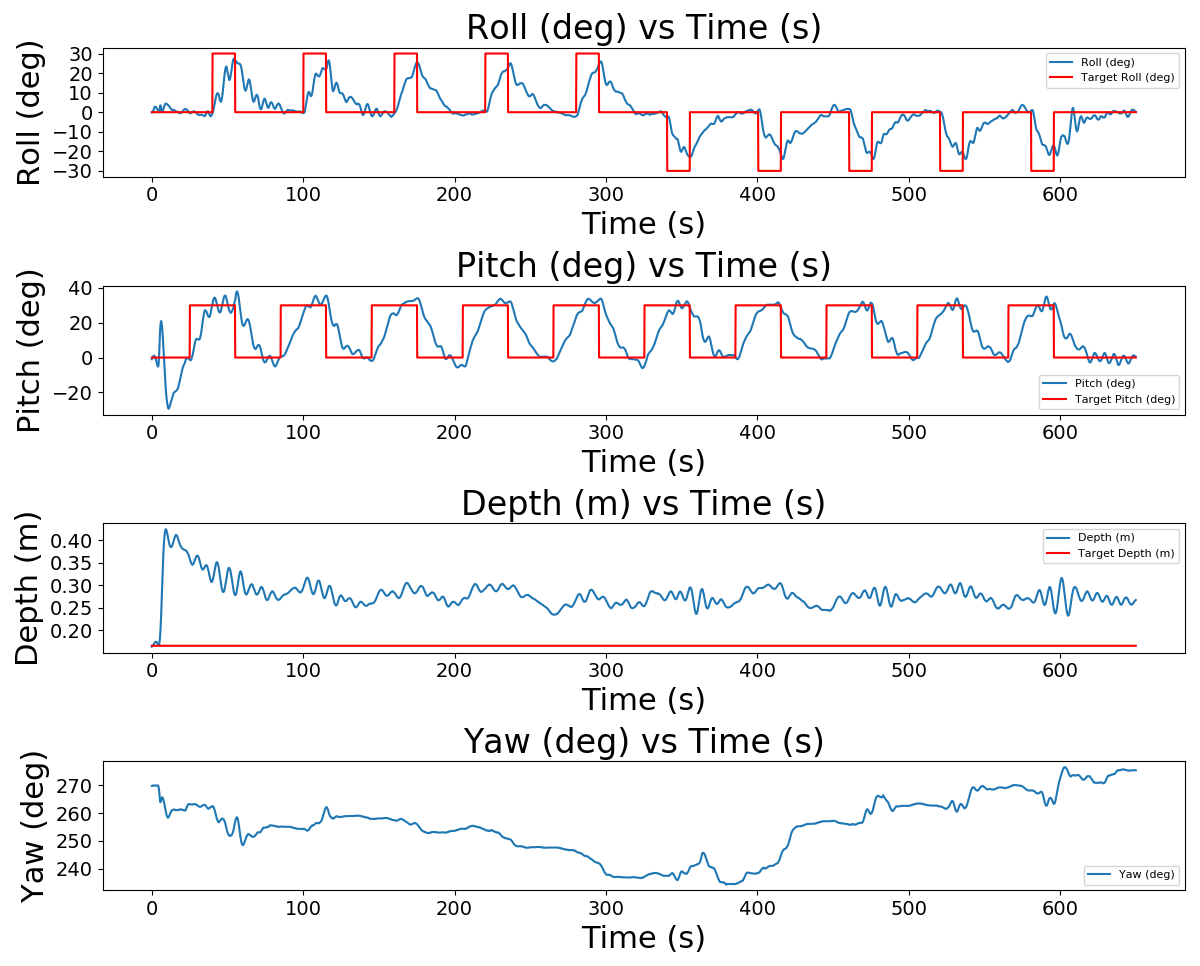}
       \caption{}
        \label{fig:yaw_senor_data}
    \end{subfigure}
    
    \caption{(a) Yaw experiment ground-truth: pose data of yaw demonstration gathered by an upward-facing GoPro in the bottom of the test tank recording an AprilTag attached to the bottom of the ReefGlider. Yaw is plotted in the bottom right. (b) Yaw experiment sensor feedback: roll, pitch, depth, and yaw sensor data (in blue) with their corresponding set points plotted (in red). Yaw is plotted in the bottom plot. Note that the difference in sign from \cref{fig:yaw-pose} is due to the difference in reference frame between the IMU and the AprilTag. This was an open-loop experiment, so no target yaw was provided.}
    \label{fig:yaw}
\end{figure}

Yaw can be achieved by a series of pitch-roll-pitch maneuvers. In theory, the robot can set its yaw to any arbitrary direction in one maneuver. However, this requires pitching by 90 degrees, which can induce singularities in the control. Instead, ReefGlider uses a series of smaller yaw maneuvers to get to the target yaw state. The robot can achieve this yaw maneuver with minimal lateral or depth motion. This can be seen from the external AprilTag data in \cref{fig:yaw-pose} and from the on-board sensor data in \cref{fig:yaw_senor_data}. 

\section{Discussion}
\label{sec:discussion}
In general, the experiments in \cref{sec:experiments} demonstrate that the ReefGlider is capable of sub-meter scale motions in 3-dimensions, in both the positive and negative directions along each axis, which is not possible in traditional gliders. These experiments also demonstrate zero-radius turning. However, comparing \cref{fig:pitch-sawtooth} and \cref{fig:roll-sawtooth}, we see that the vehicle is still not symmetric in terms of motion performance, as the hydrodynamic drag in the y-direction of motion is more significant than the x-direction, which is also more streamlined due to the orientation of the cylinders. 

The settling times of the PID controllers in all experiments could be improved by more accurate modelling of the nonlinear dynamics, further controller tuning, or potentially shifting to cascading or rate-based architectures.

For deployment in ocean currents, a hydrodynamic shell and a \textit{doppler velocity log} (DVL) can be added to ReefGlider. The hydrodynamic shell would introduce control surfaces that can increase performance through decreasing drag and increasing lift. The DVL can be integrated into the feedback controller, allowing ReefGlider to sense and counter-act or leverage currents by orienting itself advantageously. 

Additionally, the current robot has an estimated maximum depth of 12 meters based on hydrostatic pressure and the maximum dynamic load of the actuators. 
In terms of physical components, we could improve the physical layout by selecting a more symmetric (potentially radially) passive component and the orientation of the buoyancy engines. 

Finally, ReefGlider can accommodate a wide range of sensor payloads, such as CTD probes, cameras, and hydrophones. While certain payloads, such as water samplers, may cause large changes in hydrodynamics, buoyancy, and mass, most can be countered by including additional passive flotation or weight. However, two considerations are the (1) relative shift in location of the center of mass and buoyancy and (2) impact of geometry of the payload on the overall hydrodynamics (mostly drag). One would need to verify that the ReefGlider has sufficient controllable buoyancy to counteract these modified terms. Alternatively, ReefGlider's VBC system can be used as a payload on other AUVs to provide buoyancy control.

\section{Conclusion}
\label{sec:conclusion}

In this paper we have presented ReefGlider, a novel robotic platform that implements vectored buoyancy control via multiple buoyancy engines to enable higher-maneuverability, when compared to existing underwater glider platforms. The robot has low noise signature and low power consumption, which makes it well suited for long-term monitoring tasks in complex and dynamic marine environments such as coral reefs. We provide a general design framework that includes hardware design and initial control implementation, as well as real-world experiments that demonstrate the maneuverability of the system. Future developments will include improved non-linear control and long-horizon planning to deal with the non-holonomic nature of the platform, better characterization of the noise and energy profiles, and finally deployments in real coral reef environments.

\section*{Acknowledgment}
This project was funded in part by the NSF NRI Award No. 2133029 and WHOI’s Ocean Vision Program. We thank K. VandenLangenberg, J. Schickert, P. Shields for providing project support including prototyping, and P. Alicandro for photo editing help.

\bibliographystyle{IEEEtran}
\bibliography{refs}

\vspace{12pt}

\end{document}